\pdfoutput=1

\documentclass[11pt]{article}

\usepackage[]{acl}

\usepackage{times}
\usepackage{latexsym}

\usepackage[T1]{fontenc}

\usepackage[utf8]{inputenc}

\usepackage{microtype}
\usepackage{tabularx}
\usepackage{multirow}
\usepackage{booktabs}
\usepackage{graphicx}
\usepackage{ragged2e}
\usepackage{footnote}
\usepackage{bbm}
\usepackage{bm}
\usepackage{latexsym}
\usepackage{amsmath}
\usepackage{amssymb}

\newcommand{\exteval}{\textsc{ExtEval}}

\newcommand{\incorcoref}{\textsc{IncorCorefEval}}
\newcommand{\incomcoref}{\textsc{IncomCorefEval}}

\newcommand{\incomdisco}{\textsc{IncomDiscoEval}}
\newcommand{\sentibias}{\textsc{SentiBias}}

\title{Extractive is not Faithful: An Investigation of Broad Unfaithfulness \\ Problems  in Extractive Summarization}

\author{Shiyue Zhang\thanks{\,\, Equal contribution.} $\;\;\;\;$ David Wan\footnotemark[1] $\;\;\;\;$ Mohit Bansal \\
  UNC Chapel Hill  \\
  {\tt \{shiyue, davidwan, mbansal\}@cs.unc.edu} 
}

\begin{document}
\maketitle
\begin{abstract}
The problems of unfaithful summaries have been widely discussed under the context of abstractive summarization. Though extractive summarization is less prone to the common unfaithfulness issues of abstractive summaries, 
does that mean \emph{extractive} is equal to \emph{faithful}? Turns out that the answer is \emph{no}.
In this work, we define a typology with
five types of broad unfaithfulness problems (including and beyond not-entailment) that can appear in extractive summaries, including \emph{incorrect coreference}, \emph{incomplete coreference}, \emph{incorrect discourse}, \emph{incomplete discourse}, as well as \emph{other misleading information}.
We ask humans to label these problems out of 1600 English summaries produced by 16 diverse extractive systems.
We find that 30\% of the summaries 
have at least one of the five issues.
To automatically detect these problems, we find that 5 existing faithfulness evaluation metrics for summarization have poor correlations with human judgment. To remedy this, we propose a new metric, \emph{\exteval}, 
that is designed for detecting unfaithful extractive summaries and is shown to have the best performance.
We hope our work can increase the awareness of unfaithfulness problems in extractive summarization and help future work to evaluate and resolve these issues.\footnote{Our data and code are publicly available at \url{https://github.com/ZhangShiyue/extractive_is_not_faithful}.}
\end{abstract}

\section{Introduction}

Text summarization is the process of distilling the most important information from a source to produce an abridged version for a particular user or task~\cite{maybury1999advances}. 
Although there are many types of text summarization tasks, in this work, we focus on the task of \emph{general purpose single document summarization}.
To produce summaries, usually either \emph{extractive summarization} methods, i.e., extracting sentences from the source, or \emph{abstractive summarization} methods, i.e., generating novel text, are applied~\cite{saggion2013automatic}. 

Abstractive summarization attracts more attention from recent works because it can produce more coherent summaries and behaves more like humans \cite{cohn-lapata-2008-sentence}.  
Impressive progress has been made for abstractive summarization by large-scale pre-trained models \cite{lewis-etal-2020-bart, zhang2020pegasus}. However, unfaithfulness problems, i.e., hallucinating new information or generating content that contradicts the source, are widely spread across models and tasks~\cite{cao2018faithful, maynez-etal-2020-faithfulness}.  Although these problems do not necessarily get captured by typically-used evaluation metrics, e.g., ROUGE \cite{lin-2004-rouge}, even minor unfaithfulness can be catastrophic and drive users away from real-world applications. 
Therefore, an increasing volume of research has focused on analyzing \cite{falke-etal-2019-ranking, maynez-etal-2020-faithfulness, goyal-durrett-2021-annotating}, evaluating \cite{kryscinski-etal-2020-evaluating, goyal-durrett-2021-annotating, wang-etal-2020-asking, durmus-etal-2020-feqa, scialom-etal-2021-questeval, xie-etal-2021-factual-consistency}, or addressing \cite{cao2018faithful, li-etal-2018-ensure, fanrobust2018, chen-etal-2021-improving, cao-wang-2021-cliff, xu2021sequence, wan-bansal-2022-factpegasus} unfaithfulness problems in abstractive summarization. 

Extractive summarization is known to be faster, more interpretable, and more reliable \cite{chen-bansal-2018-fast, li-etal-2021-ease, dreyer2021analyzing}. And the selection of important information is the first skill that humans learn for summarization \cite{kintsch1978cognitive, brown1983macrorules}. 
Recently, some works discuss the trade-off between abstractiveness and faithfulness \cite{ladhak-etal-2022-faithful, dreyer2021analyzing}. They find that the more extractive the summary is, the more faithful it is.\footnote{Note that some previous works seemed to interchange the usage of \emph{factuality} and \emph{faithfulness}. But we think they are slightly different. Thus, we
stick to \emph{faithfulness} that represents the property of staying true to the source.}
This may give the community the impression that if the content is extracted from the source, it is guaranteed to be faithful. However, is this always true? In this work, we will show that, unfortunately, it is not.

The problems of extractive summarization are usually referred as \emph{coherence}, \emph{out-of-context}, or \emph{readability} issues \cite{nanbaokumura2000producing, nenkova2012survey, saggion2013automatic, dreyer2021analyzing}. Though they may sound irrelevant to faithfulness, 
some early works give hints of their unfaithful ingredients. \citet{gupta2010survey} describe the `dangling' anaphora problem -- sentences often contain pronouns that lose their referents when extracted out of context, and stitching together extracts may lead to \emph{a misleading interpretation of anaphors}. 
\citet{barzilay-etal-1999-information} comment on extractive methods for multi-document summarization, that extracting some similar sentences could produce \emph{a summary biases towards some sources}. \citet{cheung2008comparing} says
that sentence extraction produces extremely incoherent text that \emph{did not seem to convey the gist of the overall controversiality} of the source.
These all suggest that even though all information is extracted directly from the source, the summary is not necessarily \emph{faithful} to the source. However, none of these works has proposed an error typology nor quantitatively answered how unfaithful the model extracted summaries are, which motivates us to fill in this missing piece. 

In this work, we conduct a thorough investigation of the broad unfaithfulness problems in extractive summarization.
Although the literature of abstractive summarization usually limits unfaithful summaries to those that are \emph{not entailed} by the source \cite{maynez-etal-2020-faithfulness, kryscinski-etal-2020-evaluating}, we discuss \emph{broader unfaithfulness} issues including and beyond
not-entailment.  
We first
design a typology consisting
five types of unfaithfulness problems that could happen in extractive summaries: \emph{incorrect coreference}, \emph{incomplete coreference}, \emph{incorrect discourse}, \emph{incomplete discourse}, and \emph{other misleading information} (see definitions in Figure~\ref{fig:type}). Among them, \emph{incorrect coreference}  and \emph{incorrect discourse} are not-entailment based errors. An example of incorrect coreference is shown in Summary 1 of Figure~\ref{fig:unfaithfulness-problems-example},
where \emph{that} in the second sentence should refer to the second document sentence --\emph{But they do leave their trash}, but it incorrectly refers to the first sentence in the summary.
Summaries with \emph{incomplete coreferences or discourses} are usually entailed by the source, but they can still lead to unfaithful interpretations.
Lastly, inspired by \emph{misinformation} \cite{o2019misinformation}, our misleading information error type refers to other cases where, despite being entailed by the source, the summary still misleads the audience by selecting biased information, giving the readers wrong impressions, etc  (see Section~\ref{sec:problem}).

\begin{figure*}[ht]
\begin{center}
\small
\resizebox{0.98\textwidth}{!}{%
\begin{tabularx}{\textwidth}{X}
\toprule 
\textbf{Document:} \\
\underline{(CNN) Most climbers who try don't succeed in summiting the 29,035-foot-high Mount Everest, the world's tallest peak.} \\ \underline{\textcolor{blue}{But they do leave their trash. Thousands of pounds of it.}} \\ \underline{\textcolor{blue}{That}'s why an experienced climbing group from the Indian army plans to trek up the 8,850-meter mountain to pick up at} \underline{least 4,000 kilograms (more than 8,000 pounds) of waste from the high-altitude camps, according to India Today.} \\ \underline{The mountain is part of the Himalaya mountain range on the border between Nepal and the Tibet region.} \\ \underline{The 34-member team plans to depart for Kathmandu on Saturday and start the ascent in mid-May.} \\ \underline{The upcoming trip marks the 50th anniversary of the first Indian team to scale Mount Everest} [...] \\ 
\underline{More than 200 climbers have died} attempting to climb the peak, part of a UNESCO World Heritage Site. \\ The Indian expedition isn't the first attempt \underline{to clean up the trash left by generations of hikers}[...] \\
\midrule
\textbf{Summary 1 (\emph{incorrect coreference}):} \\
(CNN) \textcolor{red}{Most climbers who try don't succeed in summiting the 29,035-foot-high Mount Everest, the world's tallest peak.}\\
\textcolor{red}{\bf That}'s why an experienced climbing group from the Indian army plans to trek up the 8,850-meter mountain to pick up at least 4,000 kilograms (more than 8,000 pounds) of waste from the high-altitude camps, according to India Today. [...] \\
\midrule
\textbf{Summary 2 (\emph{incomplete coreference \& incorrect discourse}) :} \\
\textbf{That}'s why an experienced climbing group from the Indian army plans to trek up the 8,850-meter mountain \\
to pick up at least 4,000 kilograms \\
\textbf{More than 200 climbers have died} \\
\textbf{to clean up the trash} [...] \\
\midrule
\textbf{Summary 3 (\emph{incomplete discourse \& incomplete coreference}):} \\
\textbf{But} \textbf{they} do leave their trash. Thousands of pounds of it. [...] \\
\bottomrule
\end{tabularx}
}
\end{center}
\caption{An example from CNN/DM~\cite{hermann2015teaching} testing set showing the first four types of unfaithfulness problems defined in section \ref{sec:problem}. The three summaries are generated by NeuSumm \cite{zhou-etal-2018-neural} Oracle (disco) \cite{xu-etal-2020-discourse}, and BERT+LSTM+PN+RL \cite{zhong-etal-2019-searching}, respectively. All extracted sentences or discouse units are \underline{underlined} in the document. The problematic parts are \textbf{bolded} in the summary. The incorrect reference in the summary is marked with \textcolor{red}{red}, and the correct reference is marked with \textcolor{blue}{blue} in the document. We replace non-relevant sentences with [...].
}
\label{fig:unfaithfulness-problems-example}
\end{figure*}

We ask humans to label these problems out of 1600 model extracted summaries that are produced by 16 extractive summarization systems 
for 100 CNN/DM English articles \cite{hermann2015teaching}. 
These 16 systems cover both supervised and unsupervised methods, include both recent neural-based and early graph-based models, and extract sentences or elementary discourse units (see Section~\ref{sec:human}). By analyzing human annotations, we find that 30.3\% of the 1600 summaries have at least one of the five types of errors.
Out of which, 3.9\% and 15.4\% summaries contain incorrect and incomplete coreferences respectively, 1.1\% and 10.7\%  summaries have incorrect and incomplete discourses respectively, and other 4.9\% summaries still mislead the audience without having coreference or discourse issues.
The non-negligible error rate demonstrates that extractive is not necessarily faithful.
Among the 16 systems, we find that the two oracle extractive systems
(that maximize ROUGE \cite{lin-2004-rouge} against the gold summary by using extracted discourse units or sentences)
surprisingly have the most number of problems, 
while the Lead3 model (the first three sentences of the source document) causes the least number of issues. 

We examine whether these problems can be automatically detected by 5 widely-used metrics, including ROUGE \cite{lin-2004-rouge} and 4 faithfulness evaluation metrics for abstractive summarization (FactCC~\cite{kryscinski-etal-2020-evaluating}, DAE~\cite{goyal-durrett-2020-evaluating}, QuestEval~\cite{scialom-etal-2021-questeval}, BERTScore~\cite{zhang2020bertscore}). We find that, except BERTScore, they have either no or small correlations with human labels.  
We design a new metric, \exteval{}, for extractive summarization. It contains four sub-metrics that are used to detect incorrect coreference, incomplete coreference, incorrect or incomplete discourse, and sentiment bias, respectively.
We show that \exteval{} performs best at detecting unfaithful extractive summaries (see Section~\ref{sec:auto-eval} for more details). 
Finally, we discuss 
the generalizability and future directions of our work in Section~\ref{sec:discussion}.

In summary, our contributions are (1) a taxonomy of broad unfaithfulness problems in extractive summarization, (2) a human-labeled evaluation set with 1600 examples
from 16 diverse extractive systems,
(3) meta-evaluations of 5 existing metrics, (4) a new faithfulness metric (\exteval{}) for extractive summarization. Overall, we want to remind the community that even when the content is extracted
from the source, there is still a chance to be unfaithful. Hence, we should be aware of these problems, be able to detect them, and eventually resolve them to achieve a more reliable summarization.

\section{Broad Unfaithfulness Problems}
\label{sec:problem}

\begin{figure*}
\centering
\small
\resizebox{0.98\textwidth}{!}{%
\begin{tabular}{p{0.1\textwidth}|p{0.6\textwidth}|p{0.21\textwidth}}
\toprule
\textbf{Type} & \textbf{Definition} & \textbf{Rationale}\\
\midrule
Incorrect Coreference & 
An anaphor in the summary refers to a different entity from what the same anaphor refers to in the document. The anaphor can be a pronoun (\emph{they}, \emph{she}, \emph{he}, \emph{it}, \emph{this}, \emph{that}, \emph{those}, \emph{these}, \emph{them}, \emph{her}, \emph{him}, \emph{their}, \emph{her}, \emph{his}, etc.) or a `determiner (\emph{the}, \emph{this}, \emph{that}, \emph{these}, \emph{those}, \emph{both}, etc.) + noun' phrase.  &  Not-entailment \\
\midrule
Incomplete Coreference & An anaphor in the summary has ambiguous or no antecedent.  & Ambiguous interpretation \\
\midrule
Incorrect Discourse & A sentence with a discourse linking term (e.g., but, and, also, on one side, meanwhile, etc.) or a discourse unit (usually appears as a sub-sentence) falsely connects to the following or preceding context in the summary, which leads the audience to infer a non-exiting fact, relation, etc. & Not-entailment \\
\midrule
Incomplete Discourse & A sentence with a discourse linking term  or a discourse unit has no necessary following or preceding context to complete the discourse. & Ambiguous interpretation \\
\midrule
Other$\;$ Misleading Information & Other misleading problems include but do not limit to leading the audience to expect a different consequence and conveying a dramatically different sentiment. & Bias and wrong impression  \\
\bottomrule
\end{tabular}
}
\caption{Our \textbf{typology} of broad unfaithfulness problems in extractive summarization.}
\label{fig:type}
\vspace{-12pt}
\end{figure*}

In this section, we will describe the five types of broad unfaithfulness problems (Figure~\ref{fig:type}) we identified for extractive summarization under our typology.
In previous works about abstractive summarization, \emph{unfaithfulness} usually only refers to the summary being \emph{not entailed} by the source \cite{maynez-etal-2020-faithfulness, kryscinski-etal-2020-evaluating}. 
The formal definition of entailment is  $t$ entails $h$ if, typically, a human reading $t$ would infer that $h$ is most likely true \cite{dagan2005pascal}. 
While we also consider being \emph{not entailed} as one of the unfaithfulness problems, we will show that there is still a chance to be unfaithful despite being entailed by the source. 
Hence, we call the five error types we define here the `broad' unfaithfulness problems, and we provide a rationale for each error type in Figure~\ref{fig:type}. 

The most frequent unfaithfulness problem of abstractive summarization is the presence of incorrect entities or predicates \cite{gabriel-etal-2021-go, pagnoni-etal-2021-understanding}, which can never happen within extracted sentences (or elementary discourse units\footnote{Elementary Discourse Unit (or EDU) is a concept from the Rhetorical Structure Theory \cite{mann1988rhetorical}. Each unit usually appears as a sub-sentence.}). For extractive summarization, the problems can only happen `across' sentences (or units).\footnote{Even though some may argue that extracted sentences should be read independently, in this work, we take them as a whole and follow their original order in the document. We think this is a reasonable assumption and shares the same spirit of previous works that talk about the coherence issue of extractive summaries \cite{gupta2010survey}.} Hence, we first define four error types about \emph{coreference} and \emph{discourse}. Following SemEval-2010 \cite{semeval_2010}, we define coreference as the mention of the same textual references to an object in the discourse model, and we focus primarily on \emph{anaphors} that require finding the correct antecedent.
We ground our discourse analysis for systems that extract sentences in the Penn Discourse Treebank \cite{prasad-etal-2008-penn}, which considers the discourse relation between sentences as ``lexically grounded''. E.g., the relations can be triggered by subordinating conjunctions (\emph{because}, \emph{when}, etc.), coordinating conjunctions (\emph{and}, \emph{but}, etc.), and discourse adverbials (\emph{however}, \emph{as a result}, etc). We refer to such words as \emph{discourse linking terms}. For systems that extract discourse units, we follow the Rhetorical Structure Theory \cite{mann1988rhetorical} and assume every unit potentially requires another unit to complete the discourse. 

Finally, inspired by the concept of \emph{misinformation} (incorrect or misleading information presented as fact), we define the fifth error type -- \emph{misleading information} that captures other misleading problems besides the other four errors. The detailed definitions of the five error types are as follows:

\textbf{Incorrect coreference} happens when the same anaphor is referred to different entities given the summary and the document. The anaphor can be a pronoun (\emph{they}, \emph{she}, \emph{he}, \emph{it}, etc.) or a `determiner (\emph{the}, \emph{this}, \emph{that}, etc.) + noun' phrase.
This error makes the summary not entailed by the source.
An example is Summary 1 of Figure~\ref{fig:unfaithfulness-problems-example},
where the mention \emph{that} refers to the sentence --\emph{But they do leave their trash. Thousands of pounds of it} -- in the document but incorrectly refers to \emph{Most climbers who try don’t succeed in summiting the 29,035-foot-high Mount Everest}. Users who only read the summary may think there is some connection between cleaning up trash and the fact that most climbers do not succeed in summiting the Mount Everest. 

\textbf{Incomplete coreference} happens when an anaphor in the summary has ambiguous or no antecedent.\footnote{\label{footnote:incomcoref}Note that sometimes a `determiner + noun' phrase has no antecedent, but it does not affect the understanding of the summary or there is no antecedent of the mention in the document either. In which case, it is \emph{not} an anaphor, and thus we do \emph{not} consider it as an incomplete coreference.} Following the formal definition of entailment, these examples are considered to be entailed by the document. Nonetheless,
it sometimes can still cause unfaithfulness, as it leads to `ambiguous interpretation'. For example, given the source ``Jack eats an orange. John eats an apple'' and the faithfulness of ``He eats an apple'' depends entirely on whom ``he'' is. 
Figure~\ref{fig:unfaithfulness-problems-example} illustrates an example of incomplete coreference, where
Summary 2 starts with \emph{that's why}, but readers of that summary do not know the actual reason. 
Please refer to Figure~\ref{fig:coreference-example} in the Appendix for another example with a dangling pronoun and ambiguous antecedents.

\textbf{Incorrect discourse} happens when a sentence with a discourse linking term (e.g., but, and, also, etc.)\footnote{We do not consider implicit (without a linking term) discourse relations between sentences because it hardly appears in our data and will cause a lot of annotation ambiguity.} or a discourse unit 
falsely connects to the following or preceding context in the summary, which leads the audience to infer a non-exiting fact, relation, etc. An example 
is shown by Summary 2 in Figure~\ref{fig:unfaithfulness-problems-example}, where \emph{More than 200 climbers have died} falsely connects \emph{to clean up the trash}, which makes readers believe 200 climbers have died because of cleaning up the trash. But in fact, they died attempting to climb the peak. This summary is also clearly not entailed by the source. 

\textbf{Incomplete discourse} happens when a sentence with a discourse linking term or a discourse unit has no necessary following or preceding context to complete the discourse.
Similar to incomplete coreference, summaries with this error are considered entailed, but the broken discourse
makes the summary 
confusing and thus
may lead to problematic interpretations.
An example is shown in Figure~\ref{fig:unfaithfulness-problems-example}.
Summary 3 starts with \emph{but}, and readers expect to know what leads to this turning, but it is never mentioned. See Figure~\ref{fig:discourse-example} for another example 
that may leave readers with a wrong impression because of incomplete discourse.

\textbf{Other misleading information} refers to other misleading problems besides the other four error types. It includes but does not limit to leading the audience to expect a different consequence and conveying a dramatically different sentiment. This error is also difficult to capture using the entailment-based definition. 
Summaries always select partial content from the source, however, sometimes, the selection can mislead or bias the audience. \citet{gentzkow2015media} show that filtering and selection can result in `media bias'. 
We define this error type so that annotators can freely express
whether they think the summary has some bias or leaves them with a wrong impression. 
The summary in Figure~\ref{fig:misleading-example} is labeled as misleading by two annotators because it can mislead the audience to believe that the football players and pro wrestlers won the contest and ate 13 pounds of steak.

Note that we think it is also valid to separate misleading information and incomplete coreference/discourse, as they are \emph{less} severe in unfaithfulness compared to not-entailment-based incorrect coreference/discourse, but we choose to cover all of them under the `broad unfaithfulness’ umbrella for completeness.

\section{Human Evaluation}
\label{sec:human}
In this section, we describe how we ask humans to
find and annotate
the unfaithfulness problems. 

\subsection{Data}
We randomly select 100 articles from CNN/DM test set \cite{hermann2015teaching} because it is a widely used benchmark for single-document English summarization and extractive methods perform decently well on it. The dataset is distributed under an Apache 2.0 license.\footnote{\url{https://huggingface.co/datasets/cnn_dailymail}} We use 16 extractive systems to produce summaries, i.e., 1600 summaries in total. 
We retain the order of sentences or units in the document as their order in the summary. 

\textbf{Ten supervised systems:} 
(1) \textbf{Oracle} maximizes the ROUGE between the extracted summary and the ground-truth summary; (2) \textbf{Oracle (discourse) \cite{xu-etal-2020-discourse}} extracts discourse units instead of sentences to maximize ROUGE while considering discourse constraints; (3) \textbf{RNN Ext RL} \cite{chen-bansal-2018-fast}; (4) \textbf{BanditSumm} \cite{dong-etal-2018-banditsum}; (5) \textbf{NeuSumm} \cite{zhou-etal-2018-neural-document}; (6) \textbf{Refresh} \cite{narayan-etal-2018-ranking}; (7) \textbf{BERT+LSTM+PN+RL} \cite{zhong-etal-2019-searching}; (8) \textbf{MatchSumm} \cite{zhong-etal-2020-extractive}; (9) \textbf{HeterGraph} \cite{wang-etal-2020-heterogeneous}; (10) \textbf{Histruct+} \cite{ruan-etal-2022-histruct}. We implement the Oracle system, and we use the open-sourced code of RNN Ext RL\footnote{\href{https://github.com/ChenRocks/fast\_abs\_rl}{https://github.com/ChenRocks/fast\_abs\_rl}} and output of Oracle (discourse)\footnote{\href{https://github.com/jiacheng-xu/DiscoBERT}{https://github.com/jiacheng-xu/DiscoBERT}}. We get summaries from Histruct+ using their released code and model.\footnote{\href{https://github.com/QianRuan/histruct}{https://github.com/QianRuan/histruct}}
The summaries of other systems are
from REALSumm \cite{bhandari-etal-2020-evaluating} open-sourced data.\footnote{\href{https://github.com/neulab/REALSumm}{https://github.com/neulab/REALSumm}}

\textbf{Six unsupervised systems:} 
(1) \textbf{Lead3} extracts the first three sentences of the document as the summary; (2) \textbf{Textrank} \cite{mihalcea-tarau-2004-textrank}; (3) \textbf{Textrank (ST)}: ST stands for Sentence Transformers \cite{reimers-gurevych-2019-sentence}; (4) \textbf{PacSum (tfidf)} and (5) \textbf{PacSum (bert)} \cite{zheng-lapata-2019-sentence}; (6) \textbf{MI-unsup} \cite{padmakumar-he-2021-unsupervised}. We implement Lead3 and use the released code of PacSum.\footnote{\href{https://github.com/mswellhao/PacSum}{https://github.com/mswellhao/PacSum}} For Textrank, we use the \texttt{summa} package.\footnote{\href{https://github.com/summanlp/textrank}{https://github.com/summanlp/textrank}} For MI-unsup, we directly use the system outputs open-sourced by the authors.\footnote{\href{https://github.com/vishakhpk/mi-unsup-summ}{https://github.com/vishakhpk/mi-unsup-summ}}

Even though only Oracle (discourse) explicitly uses the discourse structure (the Rhetorical Structure Theory graph), some other systems also implicitly model discourse, e.g., HeterGraph builds a graph of sentences based on word overlap. 

\begin{figure}
    \centering
    \includegraphics[width=0.48\textwidth]{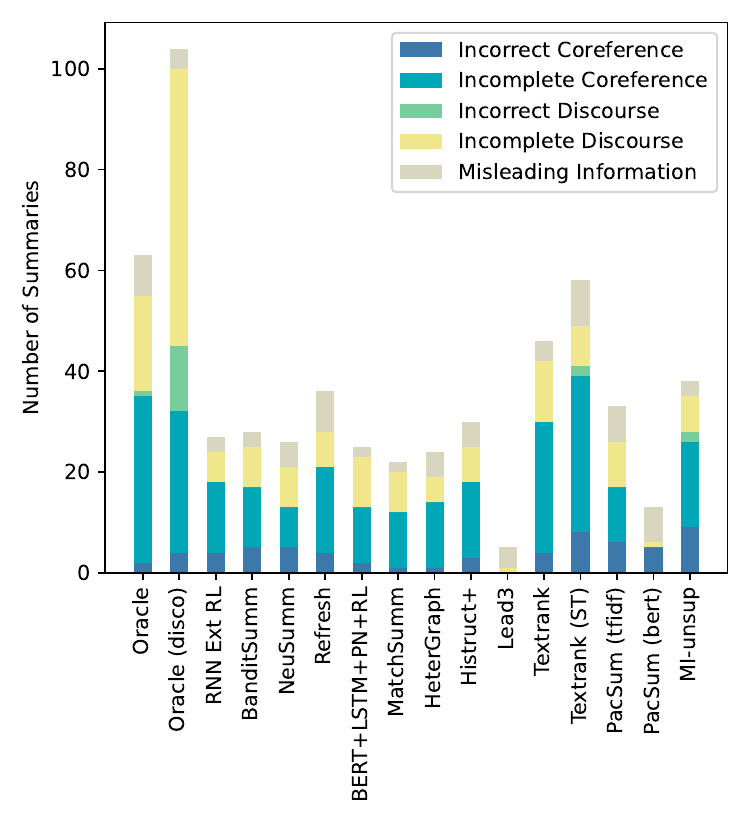}
    \vspace{-15pt}
    \caption{The unfaithfulness error distributions of 16 extractive summarization systems.}
    \label{fig:error}
\end{figure}

\subsection{Setup} 
\label{sec:human-setup}
We ask humans to label unfaithfulness problems out of the 1600 system summaries. The annotation interface (HTML page) is shown in Figure~\ref{fig:interface} in the Appendix. It first shows the summary and the document. The summary sentences are also underlined in the document. To help with annotation, we run a state-of-the-art coreference resolution model, SpanBERT \cite{joshi-etal-2020-spanbert} via AllenNLP (v2.4.0)  \cite{gardner2018allennlp}
on the summary and the document respectively. Then, mentions from the same coreference cluster will be shown in the same color. Since the coreference model can make mistakes, we ask annotators to use them with caution.   

Annotators are asked to judge whether the summary has each of the five types of unfaithfulness via five \emph{yes or no} questions
 and if yes, justify the choice by pointing out the unfaithful parts. Details of the annotation can be found in Appendix~\ref{sec:appendix_human_eval}.

Four annotators, two of the authors (PhD students trained in NLP/CL) and two other CS undergraduate students (researchers in NLP/CL), conducted all annotations carefully in about 3 months. Each of the 1600 summaries first was labeled by two annotators independently. Then, they worked together to resolve their differences in annotating incorrect/incomplete coreferences and incorrect/incomplete discourses because these errors have little subjectivity and agreements can be achieved. The judgment of misleading information is more subjective. Hence, each annotator independently double-checked examples that they labeled \emph{no} while their partner labeled \emph{yes}, with their partner's answers shown to them. They do not have to change their mind if they do not agree with their partner. This step is meant to make sure nothing is missed by accident.
In total, 149 examples have at least one misleading label, out of which, 79 examples have both annotators' misleading labels. In analysis, we only view a summary as misleading when both annotators labeled \emph{yes}, regardless of the fact that they may have different reasons. 

\subsection{Results of Human Evaluation}
\label{sec:human-eval}
Finally, we find that 484 out of 1600 (30.3\%) summaries contain at least one of the five problems. 63 (3.9\%) summaries contain incorrect coreferences, 247 (15.4\%) summaries have incomplete coreferences, 18 (1.1\%) summaries have incorrect discourses, 171 (10.7\%) have incomplete discourses,  and 79 (4.9\%) summaries are misleading. The error breakdowns for each system are illustrated in Figure~\ref{fig:error}. Note that one summary can have multiple problems, hence why Oracle (discourse) in Figure~\ref{fig:error} has more than 100 errors.

The nature of different models makes them have different chances to create unfaithfulness problems. For example, the Lead3 system has the least number of problems because the first three sentences of the document usually have an intact discourse, except in a few cases it requires one more sentence to complete the discourse.
In contrast, the two Oracle systems have the most problems. The Oracle model often extracts sentences from the middle part of the document, i.e., having a higher chance to cause dangling anaphora or discourse linking. The Oracle (discourse) model contains the most number of incorrect discourses because concatenating element discourse units together increases the risk of misleading context. Furthermore, better systems w.r.t ROUGE scores do not necessarily mean that the summaries are more faithful; the latest system Histruct+ still contains many unfaithfulness errors, indicating the need to specifically address such faithfulness issues.

\citet{cao2018faithful} show that about 30\% abstractive summaries generated for CNN/DM are not entailed by the source. Also on CNN/DM, FRANK \cite{pagnoni-etal-2021-understanding} finds that about 42\% abstractive summaries are unfaithful, including both entity/predicate errors and coreference/discourse/grammar errors. Compared to these findings, extractive summarization apparently has fewer issues. We do note, however, that the quantity is not negligible, i.e., extractive $\neq$ faithful.

\section{Automatic Evaluation}
\label{sec:auto-eval}

Here, we analyze whether existing automatic faithfulness evaluation metrics can detect unfaithful extractive summaries. We additionally propose a new evaluation approach, \exteval{}.

\subsection{Meta-evaluation Method} \label{sec:meta-evel-method}
To evaluate automatic faithfulness evaluation metrics (i.e., meta-evaluation) for extractive summarization,
we follow the faithfulness evaluation literature of abstractive summarization \cite{durmus-etal-2020-feqa, wang-etal-2020-asking, pagnoni-etal-2021-understanding} and compute the correlations between metric scores and human judgment on our meta-evaluation dataset (i.e., the 1600 examples). 
Though one summary can have multiple issues for one error type, for simplicity, we use the binary (0 or 1) label as the human judgment of each error type. In addition, we introduce an \textbf{Overall} human judgment by taking the \emph{summation} of the five error types. So, the maximum score of Overall is 5.
We use Pearson $r$ or Spearman $\rho$ as the correlation measure. 

This meta-evaluation method is essentially
assessing how well the metric can automatically detect unfaithful summaries, which is practically useful. For example, we can pick out summaries with
high unfaithfulness scores and ask human editors to fix them. 
One underlying assumption is that the metric score is comparable across examples. However, some metrics are example-dependent (one example's score of 0.5 $\neq$ another example's score of 0.5), e.g., ROUGE is influenced by summary length \cite{sun-etal-2019-compare}. In practice, we do not observe any significant effect of example dependence on our correlation computation.

To understand the correlation without example-dependence issues, we provide two alternative evaluations \emph{system-level} and \emph{summary-level} correlations, which have been reported in a number of previous works \cite{peyrard-etal-2017-learning, bhandari-etal-2020-evaluating, deutsch-etal-2021-towards, zhang-bansal-2021-finding}. These two correlations assess the effectiveness of the metrics to rank systems.
We define the correlations and present the results in Appendix \ref{app:meta-eval}.

\subsection{Existing Faithfulness Evaluation Metrics}

In faithfulness evaluation literature, a number of metrics have been proposed for abstractive summarization. 
They can be roughly categorized into two groups: entailment classification and question generation/answering (QGQA). Some of them assume that the extractive method is inherently faithful.

We choose FactCC \cite{kryscinski-etal-2020-evaluating} and DAE \cite{goyal-durrett-2020-evaluating} as representative entailment classification metrics. However, since they are designed to check whether each sentence or dependency arc is entailed by the source, we suspect that they cannot detect discourse-level errors. QuestEval \cite{scialom-etal-2021-questeval} is a representative QGQA metric, which theoretically can detect \emph{incorrect coreference} because QG considers the long context of the summary and the document. We also explore BERTScore Precision \cite{zhang2020bertscore} that is shown to well correlate with human judgment of faithfulness \cite{pagnoni-etal-2021-understanding, fischer2021finding}, as well as ROUGE-2-F1 \cite{lin-2004-rouge}. Details of these metrics can be found in Appendix~\ref{sec:appendix_metrics}.

Note that for all metrics except for DAE, we \textbf{negate} their scores before computing human-metric correlations  because we want them to have higher scores when the summary is more unfaithful, just like our human labels. Table~\ref{table:system_scores} in the Appendix shows their original scores for the 16 systems.

\begin{table*}[t]
\centering
\small
\resizebox{0.98\textwidth}{!}{%
\begin{tabular*}{1.01\textwidth}{l|cc|cc|cc|cc|cc|cc}
\toprule
& \multicolumn{2}{c}{\bf Incor. Coref.} & \multicolumn{2}{c}{\bf Incom. Coref.} & \multicolumn{2}{c}{\bf Incor. Disco.} & \multicolumn{2}{c}{\bf Incom. Disco.} & \multicolumn{2}{c}{\bf Mislead.} & \multicolumn{2}{c}{\bf Overall} \\
\cmidrule(lr){2-3} \cmidrule(lr){4-5} \cmidrule(lr){6-7} \cmidrule(lr){8-9} \cmidrule(lr){10-11} \cmidrule(lr){12-13}
\bf Metrics & $r$ & $\rho$ & $r$ & $\rho$ & $r$ & $\rho$ & $r$ & $\rho$ & $r$ & $\rho$ & $r$ & $\rho$ \\
\midrule
-ROUGE-2-F1 & 0.05 & 0.06 & 0.03 & 0.08 & -0.07 & -0.07 & -0.14 & -0.10 & 0.03 & 0.03 & -0.04 & 0.02 \\
-FactCC &  -0.04 & -0.04 & 0.05 & 0.02 & \bf 0.24 & 0.17 & 0.10 & 0.03 & -0.00 & 0.01 & 0.11 & 0.05 \\
DAE & 0.01 & 0.04 & 0.04 & 0.08 & 0.02 & 0.04 & -0.01 & 0.02 & 0.06 & 0.03 & 0.05 & 0.07 \\
-QuestEval & 0.09 & 0.12 & 0.14 & 0.15 & -0.01 & 0.01 & 0.05 & 0.06 & 0.08 & 0.09 & 0.17 & 0.19 \\
-BERTScore Pre. & 0.08 & 0.09 & 0.21 & 0.20 & 0.18 & 0.15 & 0.29 & 0.25 & 0.11 & \bf 0.12 & 0.37 & 0.35 \\
\midrule
\incorcoref{} & \bf 0.25 & \bf 0.25 & 0.04 & 0.04 & -0.01 & -0.01 & -0.00 & -0.00 & 0.04 & 0.04 & 0.11 & 0.08 \\
\incomcoref{} & 0.11 & 0.11 & \bf 0.48 & \bf 0.48 & 0.06 & 0.06 & 0.16 & 0.16 & 0.01 & 0.01 & 0.42 & 0.42 \\
\incomdisco{} & 0.03 & 0.03 & 0.11 & 0.11 & 0.20 & \bf 0.20 &  \bf 0.61 &  \bf 0.61 & -0.02 & -0.02 & 0.42 & 0.38 \\
\sentibias{} & -0.02 & -0.03 & 0.07 & 0.05 & -0.01 & -0.00 & 0.09 &  0.08 &  \bf 0.14 & 0.11 & 0.13 & 0.11 \\
\exteval{} & 0.17 & 0.13 & 0.37 & 0.34 & 0.14 & 0.11 & 0.43 & 0.36 & 0.04 & 0.05 & \bf 0.54 & \bf 0.46\\
\bottomrule
 \end{tabular*}
 }
\caption{Human-metric correlations. The negative sign (-) before metrics means that their scores are negated to retain the feature that the higher the scores are the more unfaithful the summaries are.}
\label{table:meta-eval-res}
\vspace{-5pt}
\end{table*}

\subsection{A New Metric: \exteval{}}
We introduce \exteval{} that is designed for detecting unfaithful extractive summaries. Corresponding to the faithfulness problems defined in Section~\ref{sec:problem}, \exteval{} is composed of four sub-metrics described as follows. We refer the readers to Appendix~\ref{sec:appendix_exteval} for more details.

\textbf{\incorcoref{}} focuses on detecting \emph{incorrect coreferences}.
Taking advantage of the model-predicted coreference clusters by SpanBERT described in Section~\ref{sec:human-setup}, we consider the different cluster mapping of the same mention in the document and summary as \emph{incorrect coreference}.

\textbf{\incomcoref{}} detects \emph{incomplete coreferences}. We also make use of the model-predicted coreference clusters. If the first appeared mention in a summary cluster is a pronoun or `determiner + noun' phrase, and it is not the first mention in the corresponding document cluster, then the summary is considered to have an \emph{incomplete coreference}.

\textbf{\incomdisco{}} is primarily designed to detect \emph{incomplete discourse}. Concretely, we check for sentences with discourse linking terms and incomplete discourse units. We consider the summary to have a problem if a discourse linking term is present but its necessary context (the previous or next sentence) is missing or a discourse unit misses its previous unit in the same sentence. It is important to note that the detected errors also include \emph{incorrect discourse}. However, we cannot distinguish between these two errors.

\textbf{\sentibias{}} evaluates how different the summary sentiment is from the document sentiment. Sentiment bias
is easier to be quantified than other misleading problems. We use the RoBERTa-based \cite{liu2019roberta} sentiment analysis model from AllenNLP~\cite{gardner2018allennlp}\footnote{We also test sentiment analysis tools from Stanza \cite{qi-etal-2020-stanza} and Google Cloud API, but they do not work better (see Appendix~\ref{app:alter-senti-tools}).
} to get the sentiments of each sentence. We take the average of sentence sentiments as the overall sentiment of the document or the summary. Then, sentiment bias is measured by the absolute difference between summary sentiment and document sentiment.

\textbf{\exteval{}} is simply the  summation of the above sub-metrics, i.e., \textbf{\exteval{} = \incorcoref{} + \incomcoref{} + \incomdisco{} + \sentibias{}}. Same as human scores, we make \incorcoref{}, \incomcoref{}, and \incomdisco{} as binary (0 or 1) scores, while \sentibias{} is a continuous number between 0 and 1. \exteval{} corresponds to the Overall human judgment introduced in Section~\ref{sec:meta-evel-method}. Note that when one TiTAN V 12G GPU is available, it takes 0.43 seconds per example to compute \exteval{} on average.

\subsection{Meta-Evaluation Results}
Table~\ref{table:meta-eval-res} shows the human-metric correlations. First, out of the five existing metrics, BERTScore
in general works best and has small to moderate \cite{cohen1988} correlations with human judgment, FactCC has a small correlation with incorrect discourse, and other metrics have small or no correlations with human labels. Considering the fact that all these five errors can also happen in abstractive summarization, existing faithfulness evaluation metrics apparently leave these errors behind. Second, the four sub-metrics of \exteval{} (\incorcoref{}, \incomcoref{},  \incomdisco{}, and \sentibias{}) in general  demonstrate better performance than other metrics at detecting their corresponding problems.
Lastly, our \exteval{} has moderate to large \cite{cohen1988} correlations with the Overall judgment, which is greatly better than all other metrics.  

Table~\ref{table:system-level-res} in Appendix~\ref{app:meta-eval} shows the system-level and summary-level correlations. 
Our \exteval{} still has the best Pearson and Spearman correlations with the Overall score on both the system level and the summary level. 
Please see Appendix~\ref{app:meta-eval} for more discussions. 

In addition, we evaluate \exteval{} on an existing meta-evaluation benchmark, SummEval~\cite{fabbri-etal-2021-summeval}. In particular, we use a subset of SummEval that has 4 extractive systems, and we take the average of their expert-annotated consistency scores as the gold human faithfulness scores and compute its correlation with \exteval{}. We find that \exteval{} achieves the best Spearman correlations, which demonstrates the good generalizability of \exteval{}. Please refer to Appendix~\ref{app:summeval} for more details.

In summary, our \exteval{} is better at identifying unfaithful extractive summaries than the 5 existing metrics we compare to. Its four sub-metrics can be used independently to examine
the corresponding unfaithfulness problems.

\section{Generalizability \& Future Work}
\label{sec:discussion}

One future direction for resolving these unfaithfulness problems is to use the errors
automatically detected by \exteval{} as hints for humans or programs to fix the summary by doing necessary yet minimal edits.
Here we illustrate the possibility for \emph{incorrect coreference}.
We manually examined the automatically detected incorrect coreferences by \exteval{}.
28 out of 32 detected incorrect coreferences are true incorrect coreferences\footnote{It shows that \exteval{} has high precision of $87.5\%$. However, we have 60 human-labeled incorrect coreferences, so the recall is only $46.7\%$ (28 out of 60).}, which we attempt to fix by developing a simple post-edit program, similar to the revision system proposed by~\citet{nanbaokumura2000producing}. The program replaces the problematic mention in the summary with the first mention in the correct coreference cluster of the document. We manually checked the corrected examples and found that 16 out of 28 were fixed correctly (see an example in Figure~\ref{fig:post-correction-example}). We leave the improvement and the extension of post-edit systems for future work.

It is worth noting that all of the five error types we define in Section~\ref{sec:problem} can also happen in abstractive summarization, though they are less studied and measured in the literature. 
To our best knowledge, FRANK \cite{pagnoni-etal-2021-understanding} and SNaC \cite{goyal2022snac} have discussed the coreference and discourse errors in the abstractive summaries. \citet{gabriel-etal-2021-go} define a sentiment error as an adjective or adverb appearing in the summary that contradicts the source, while our misleading information has a more general definition.
We hope that our taxonomy can shed some light for future works to explore the broad unfaithfulness of all summarization methods.

\section{Conclusion}
We conducted a systematic analysis of broad unfaithfulness problems in extractive summarization. 
We proposed 5 error types
and produced a human-labeled evaluation set of 1600 examples. We found that (i) 30.3\% of the summaries have at least one of the 5 issues, (ii) existing metrics correlate poorly with human judgment, and (iii) our new faithfulness evaluation metric \exteval{} performs the best at identifying these problems. Through this work, we want to raise the awareness of unfaithfulness issues in extractive summarization and stress that \emph{extractive is not equal to faithful}.

\section*{Acknowledgments}
We thank anonymous reviewers for their valuable comments. We thank Yinuo Hu and Abhay Zala for helping with the data, and Jiacheng Xu for helping us get the system outputs of the Oracle (discourse) model. We also thank Ido Dagan for the helpful discussions. This work was supported by NSF-CAREER Award 1846185, NSF-AI Engage Institute DRL-2112635, and a Bloomberg Data Science Ph.D. Fellowship.

\section*{Limitations}
Since we focus on extractive summarization in this work, the conclusions will be more useful for summarization tasks where extractive methods perform decently well (e.g., CNN/DM \cite{hermann2015teaching}) 
compared to extremely abstractive summarization tasks
(e.g., XSum \cite{narayan-etal-2018-dont}). Experts, two of the authors (PhD students trained in NLP/CL) and two other CS undergraduate students (researchers in NLP/CL), conducted our annotations. Hence, to scale up data annotation by working with crowdsourcing workers may require additional training for the workers.
Our \exteval{} is designed for extractive summarization, which is currently not directly applicable 
for abstractive summaries except for \sentibias. 

As our data is collected on CNN/DM, the percentages of each error type may change when evaluating a different summarization dataset, though we believe that the conclusion, extractive is not faithful, will not change. To initially verify our conjecture, we manually examine 23 oracle summaries from the test set of PubMed~\cite{sen2008collective} and find 2 incorrect coreferences, 5 incomplete coreferences, 1 incorrect discourse, and 1 incomplete discourse.

\section*{Broader Impact Statement}
Many works have shown that model-generated summaries are often ``unfaithful'', where the summarization model changes the meaning of the source document or hallucinates new content~\cite{cao2018faithful, maynez-etal-2020-faithfulness}. This potentially causes misinformation in practice. Our work follows the same idea, but, as opposed to focusing on abstractive summarization, we show that even extracting content from the source document can still alter the meaning of the source document and cause misinformation. Hence, we want to remind NLP practitioners that even extractive is not faithful and these issues need to be addressed before we can trust model-produced extractive summaries for real-world applications. 

\bibliography{anthology,custom}
\bibliographystyle{acl_natbib}

\appendix

\begin{table*}[t]
\centering
\small
\resizebox{0.95\textwidth}{!}{%
\begin{tabular*}{1.02\textwidth}{l|cc|cc|cc|cc|cc|cc}
\toprule
\multicolumn{11}{l}{\bf System-level Correlations} \\
\midrule
& \multicolumn{2}{c}{\bf Incor. Coref.} & \multicolumn{2}{c}{\bf Incom. Coref.} & \multicolumn{2}{c}{\bf Incor. Disco.} & \multicolumn{2}{c}{\bf Incom. Disco.} & \multicolumn{2}{c}{\bf Mislead.} & \multicolumn{2}{c}{\bf Overall} \\
\cmidrule(lr){2-3} \cmidrule(lr){4-5} \cmidrule(lr){6-7} \cmidrule(lr){8-9} \cmidrule(lr){10-11} \cmidrule(lr){12-13}
\bf Metrics & $r$ & $\rho$ & $r$ & $\rho$ & $r$ & $\rho$ & $r$ & $\rho$ & $r$ & $\rho$ & $r$ & $\rho$ \\
\midrule
-ROUGE-2-F1 & 0.28 & \bf 0.59 & -0.39 & 0.08 & -0.78 & -0.01 & -0.88 & -0.26 & 0.01 & 0.12 & -0.71  & 0.14 \\
-FactCC &  0.29 & 0.34 & 0.44 & 0.39 & 0.81 &  0.51 & 0.81 & 0.60 & -0.13 & -0.22 & 0.75 & 0.54 \\
DAE & 0.23 & 0.26 & 0.66 & 0.39 & 0.11 & 0.41 & 0.23 & \bf 0.74 & \bf 0.64 & \bf 0.44 & 0.50 & 0.58 \\
-QuestEval & 0.27 & 0.35 & 0.16 & 0.40 & -0.26 & 0.33 & -0.25 & 0.36 & 0.18 &  0.19 & -0.06 & 0.43\\
-BERTScore Pre. & 0.29 & 0.30 & 0.50 & 0.57 & 0.70 & 0.58 & 0.73 & 0.58 & 0.09 & 0.10 & 0.74 & 0.68 \\
\midrule
\incorcoref{} & \bf 0.43 & 0.12 & 0.32 & 0.31 & -0.03 & 0.19 & -0.16 & -0.02 & 0.25 & 0.12 & 0.11 & 0.22 \\
\incomcoref{} & 0.38 & 0.34 & \bf 0.96 & \bf 0.87 & 0.52 & 0.72 & 0.59 & 0.56 & 0.20 & 0.13 & 0.85 & 0.85 \\
\incomdisco{} & 0.30 & 0.46 & 0.58 & 0.76 & \bf 0.96 & \bf 0.76 & \bf 0.92 & 0.71 & -0.06 & 0.10 & 0.90 & \bf 0.88 \\
 \sentibias{} & -0.37 & -0.48 & 0.37 & 0.18 & 0.57 & 0.19 & 0.69 &  0.32 &  0.00 & 0.01 & 0.56 & 0.09 \\
\exteval{} & 0.37 & 0.33 & 0.83 & 0.84 & 0.83 & \bf 0.76 & 0.84 & 0.67 & 0.08 & 0.09 & \bf 0.96 & \bf 0.88 \\
\bottomrule
\toprule
\multicolumn{11}{l}{\bf Summary-level Correlations} \\
\midrule
& \multicolumn{2}{c}{\bf Incor. Coref.} & \multicolumn{2}{c}{\bf Incom. Coref.} & \multicolumn{2}{c}{\bf Incor. Disco.} & \multicolumn{2}{c}{\bf Incom. Disco.} & \multicolumn{2}{c}{\bf Mislead.} & \multicolumn{2}{c}{\bf Overall} \\
\cmidrule(lr){2-3} \cmidrule(lr){4-5} \cmidrule(lr){6-7} \cmidrule(lr){8-9} \cmidrule(lr){10-11} \cmidrule(lr){12-13}
\bf Metrics & $r$ & $\rho$ & $r$ & $\rho$ & $r$ & $\rho$ & $r$ & $\rho$ & $r$ & $\rho$ & $r$ & $\rho$ \\
\midrule
-ROUGE-2-F1 & 0.09 & 0.06 & -0.05 & -0.01 & -0.47 & -0.28 & -0.37 & -0.28 & -0.00 & 0.02 & -0.22 & -0.13 \\
-FactCC &  -0.07 & -0.07 & 0.05 & 0.04 & 0.46 & 0.42 & 0.13 & 0.10 & 0.03 & 0.03 & 0.12 & 0.09 \\
DAE & 0.03 & 0.03 & 0.16 & 0.23 & 0.01 & 0.11 & 0.00 & 0.03 & \bf0.20 & \bf0.17 & 0.10 & 0.14 \\
-QuestEval & 0.10 & 0.13 & 0.17 & 0.20 & -0.13 & -0.06 & -0.03 & -0.02 & 0.06 & 0.08 & 0.08 & 0.13\\
-BERTScore Pre. & 0.11 & 0.12 & 0.24 & 0.23 & 0.48 & 0.37  & 0.36 & 0.30 & 0.10 & 0.09 & 0.36 & 0.32 \\
\midrule
\incorcoref{} & \bf 0.44 & \bf 0.44 & 0.07 & 0.07 & -0.07 & -0.07 & -0.06 & -0.06 & 0.13 & 0.13 & 0.13 & 0.12 \\
\incomcoref{} & 0.13 & 0.13 & \bf 0.52 & \bf 0.52 & 0.09 & 0.09 & 0.23 & 0.23 & 0.04 & 0.04 & 0.43 & \bf 0.43 \\
\incomdisco{} & 0.06 & 0.06 & 0.15 & 0.15 & \bf 0.65 & \bf 0.65 & \bf0.67 & \bf0.67 & -0.04 & -0.04 &  0.43 & 0.41 \\
\sentibias{} & -0.06 & -0.06 & 0.07 & 0.07 & -0.01 & 0.01 & 0.06 &  0.07 &  0.11 & 0.11 & 0.09 & 0.10 \\
\exteval{} & 0.23 & 0.16 & 0.42 & 0.37 & 0.36 & 0.28 & 0.48 & 0.37 & 0.04 & 0.07 & \bf 0.52 & \bf0.43 \\
\bottomrule
 \end{tabular*}
 }
 \vspace{-5pt}
\caption{System-level and summary-level correlations. The negative sign (-) before metrics means that their scores are negated to retain the feature that the higher the scores are the unfaithful the summaries are. } 
\label{table:system-level-res}
\end{table*}

\section*{Appendix}
\section{Another Meta-evaluation Method}
\label{app:meta-eval}

\subsection{Definitions}
\noindent\textbf{System-level} correlation evaluates \emph{how well the metric can compare different summarization systems}. 
We denote the correlation measure as $K$, human scores as $h$, the metric as $m$, and generated summaries as $s$. We assume there are $N$ documents and $S$ systems in the mete-evaluation dataset. The system-level correlation is defined as follows:
\begin{align*}
    K^{sys}_{m, h} = K(& [\frac{1}{N}\sum_{i=1}^N m(s_{i1}), ..., \frac{1}{N}\sum_{i=1}^N m(s_{iS})], \\
    & [\frac{1}{N}\sum_{i=1}^N h(s_{i1}), ..., \frac{1}{N}\sum_{i=1}^N h(s_{iS})])
\end{align*}
In our case, $N=100$ and $S=16$. We use Pearson $r$ or Spearman $\rho$ as the correlation measure $K$. 

\vspace{4pt}
\noindent\textbf{Summary-level} correlation evaluates \emph{if the metric can reliably compare summaries generated by different systems for the same document}. Using the same notations as above, it is written by:
\begin{align*}
    K^{sum}_{m, h} = \frac{1}{N}\sum_{i=1}^N K(& [m(s_{i1}), ..., m(s_{iS})], \\ & [h(s_{i1}), ..., h(s_{iS})])
\end{align*}

\begin{table*}
\centering
\small
\resizebox{0.95\textwidth}{!}{%
\begin{tabular*}{1.02\textwidth}{l|cc|cc|cc|cc|cc|cc}
\toprule
& \multicolumn{2}{c}{\bf Incor. Coref.} & \multicolumn{2}{c}{\bf Incom. Coref.} & \multicolumn{2}{c}{\bf Incor. Disco.} & \multicolumn{2}{c}{\bf Incom. Disco.} & \multicolumn{2}{c}{\bf Mislead.} & \multicolumn{2}{c}{\bf Overall} \\
\cmidrule(lr){2-3} \cmidrule(lr){4-5} \cmidrule(lr){6-7} \cmidrule(lr){8-9} \cmidrule(lr){10-11} \cmidrule(lr){12-13}
\bf Metrics & $r$ & $\rho$ & $r$ & $\rho$ & $r$ & $\rho$ & $r$ & $\rho$ & $r$ & $\rho$ & $r$ & $\rho$ \\
\midrule
\sentibias{} (AllenNLP) &  -0.02 & -0.03 & 0.07 & 0.05 & -0.01 & -0.00 & 0.09 &  0.08 &  \bf 0.15 & \bf 0.12 & 0.13 & 0.11 \\
\sentibias{} (Stanza) & 0.01 & 0.02 & -0.01 &-0.02 & 0.01 & 0.01 & 0.10 & 0.09 & 0.06 & 0.04 & 0.07 & 0.05 \\
\sentibias{} (Google) & 0.06 & 0.06 & -0.01 & -0.01 & 0.00 & 0.01 & 0.04 & 0.04  & 0.05 & 0.05 & 0.05 & 0.06 \\
\sentibias{} (ensemble) & 0.02 & 0.04 & 0.02 & 0.02 & 0.00 & -0.00 & 0.12 &  0.11  & 0.12 & 0.10  & 0.12 & 0.12\\
\bottomrule
 \end{tabular*}
 }
 \vspace{-5pt}
\caption{Comparison of using different sentiment analysis tools in SentiBias sub-metric.} 
\label{table:other-senti-tools}
\vspace{-8pt}
\end{table*}

\subsection{Results}
Table~\ref{table:system-level-res} illustrates the system-level and summary-level correlations of different metrics with human judgment. 
Note that, for both system-level and summary-level correlations, their correlations are computed between two vectors of length 16 (16 systems), whereas the meta-evaluation method we used in the main paper computes the correlations between two vectors of length 1600 (1600 examples). A smaller sample size will cause a larger variance. This is especially true for system-level correlations, because, following the definitions above, the summary-level correlation ($K^{sum}_{m, h}$) averages across N (in our case, N=100) which can help reduce the variance.

Nevertheless, as shown in Table~\ref{table:system-level-res}, our \exteval{} achieves the best Pearson and Spearman correlations with the Overall human judgment on both the system level and the summary level. 
It means \exteval{} can rank extractive systems well based on how unfaithful they are. The three sub-metrics (\incorcoref{}, \incomcoref{}, and \incomdisco{}) perform best at judging which system produces more errors of their corresponding error types. But for detecting misleading information, DAE works best.  Out of the 5 existing metrics, BERTScore Precision is the best in general, and on system level, FactCC also works decently well.

\section{Meta-evaluation Results on SummEval}
\label{app:summeval}
We mainly evaluate \exteval{} on the dataset we collected because \exteval{} is designed for detecting problematic extractive summaries and is not applicable to abstractive summaries. Nonetheless, we find a subset of SummEval~\cite{fabbri-etal-2021-summeval} that contains 4 extractive systems. We use the average of their consistency (=faithfulness) scores annotated by experts as the gold human scores and compute its correlation with \exteval{}. We apply two meta-evaluation methods: (1) Method 1, the same meta-evaluation method as Section~\ref{sec:meta-evel-method}, and (2) Method 2, the system-level evaluation introduced in \ref{app:meta-eval}, which is also used by \citet{fabbri-etal-2021-summeval}, though here we only have 4 systems. The results can be found in Table~\ref{table:summ-eval-res}. As we can observe, under both methods, our \exteval{} achieves the best Spearman correlations and competitive Pearson correlations, which demonstrates the good generalizability of \exteval{}.

\begin{table}[t]
\centering
\small
\begin{tabular*}{0.42\textwidth}{l|cc|cc}
\toprule
& \multicolumn{2}{c}{\bf Method 1} & \multicolumn{2}{c}{\bf Method 2} \\
\cmidrule(lr){2-3} \cmidrule(lr){4-5} 
\bf Metrics & $r$ & $\rho$ & $r$ & $\rho$\\
\midrule
FactCC & -0.04 & -0.11 & \bf 0.68 & 0.40\\
QuestEval & -0.04 & 0.02 & -0.46 & -0.68 \\
BERTScore Pre. & \bf 0.13 & 0.14 & -0.30 & 0.0 \\
-\exteval{} & 0.10 & \bf 0.16 & 0.31 & \bf 0.60\\
\bottomrule
 \end{tabular*}
\caption{Meta-evaluation results on SummEval~\cite{fabbri-etal-2021-summeval}. Method 1 refers to the meta-evaluation method used in Section~\ref{sec:meta-evel-method}, while Method 2 refers to the system-level correlation used by \citet{fabbri-etal-2021-summeval}. We negate \exteval{} to make higher scores mean more faithful.} 
\label{table:summ-eval-res}
\vspace{-12pt}
\end{table}

\section{Alternative Sentiment Analysis Tools}
\label{app:alter-senti-tools}
In the main paper, we use the sentiment analysis tool from AllenNLP (v2.4.0) \cite{gardner2018allennlp}\footnote{\url{https://storage.googleapis.com/allennlp-public-models/stanford-sentiment-treebank-roberta.2021-03-11.tar.gz}} to implement our \sentibias{} sub-metric of \exteval{}. Here, we test two other sentiment analysis tools from Stanza~\cite{qi-etal-2020-stanza} and Google Cloud API\footnote{\url{https://cloud.google.com/apis/docs/overview}}, respectively. We also try an ensemble method by averaging their output scores. Table~\ref{table:other-senti-tools} shows the performance. Note that these correlations are computed with 15 systems (except Histruct+) because we added Histruct+ after we conducted this analysis. Thus, the numbers are slightly different from those in Table~\ref{table:meta-eval-res}. AllenNLP works better than the other two tools. The ensemble does not help improve the performance either.

\section{Human Evaluation Details}\label{sec:appendix_human_eval}
We did not choose to label the data on Amazon Mechanical Turk because we think that understanding the concepts of coreference and discourse requires some background knowledge of linguistics and NLP. 

\begin{figure*}
    \centering
    \includegraphics[width=1.0\textwidth]{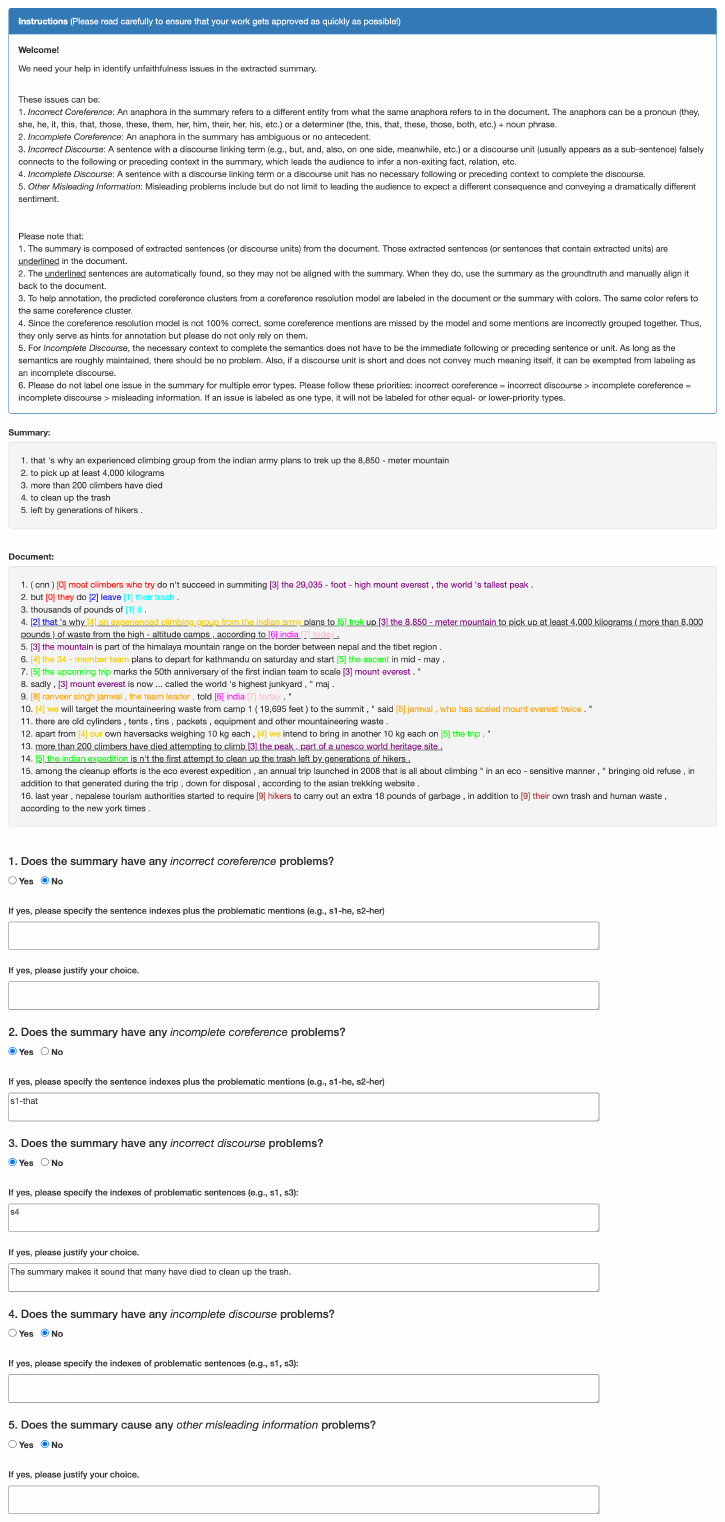}
    \caption{The interface for human annotation.}
    \label{fig:interface}
\end{figure*}

Figure~\ref{fig:interface} shows the annotation interface and an example annotation. We ask the expert annotators to justify when they think there exists an unfaithful problem. Specifically, if they think the summary has \emph{incorrect coreferences}, they need to further specify the sentence indices and the mentions. For example, ``s2-he'' means ``he'' in the second summary sentence is problematic. Meanwhile, they need to justify their answer by explaining why ``s2-he'' is an incorrect coreference. For \emph{incomplete coreference}, annotators also need to specify the sentence indices plus mentions, but no explanation is required because it can always be ``the mention has no clear antecedent.'' For \emph{incorrect discourse}, they need to specify sentence indices and justify their choice. For \emph{incomplete discourse}, they only need to specify sentence indices. We find that many summaries have multiple incomplete coreference or discourse issues. Annotators need to label all of them, separated by ``,'', e.g., ``s2-he, s3-the man''. Lastly, besides these four errors, if they think the summary can still mislead the audience, we ask them to provide an explanation to support it.

To avoid one issue in the summary being identified as multiple types of errors, we give the following priorities: incorrect coreference $=$ incorrect discourse $>$ incomplete coreference $=$ incomplete discourse $>$ other misleading information. If an issue is labeled as one type, it will not be labeled for other equal- or lower-priority types.  

\begin{table*}
\centering
\small
\resizebox{0.97\textwidth}{!}{%
\begin{tabular*}{1.05\textwidth}{l|c|c|c|c|c|c|c}
\toprule
& ROUGE-2-F1 & FactCC & DAE$\downarrow$ & QuestEval & BERTScore Pre. & \exteval{}$\downarrow$ & Human Overall$\downarrow$ \\
\midrule
Oracle & 25.09 & 0.95 & 0.02 & 0.45 & 0.92 & 0.98 & 0.63\\
Oracle (discourse) & 33.38 & 0.77 & 0.00 & 0.55 & 0.89 & 1.65 & 1.04\\
\midrule
RNN Ext RL & 12.89 & 0.97 & 0.00 & 0.49 & 0.95 & 0.59 & 0.27\\
BanditSumm & 13.48 & 0.91 & 0.00 & 0.48 & 0.93 & 0.57 & 0.28 \\
NeuSumm & 13.69 & 0.90 & 0.01 & 0.48 & 0.91 & 0.52 & 0.26 \\
Refresh & 12.96 & 0.93 & 0.00 & 0.48 & 0.92 & 0.66 & 0.36 \\
BERT+LSTM+PN+RL & 14.34 & 0.90 & 0.00 & 0.48 & 0.93 & 0.59 & 0.25\\
MatchSumm & 15.42 & 0.99 & 0.00 & 0.48 & 0.94 & 0.58 & 0.22\\
HeterGraph & 14.05 & 1.00 & 0.00 & 0.50 & 0.94 & 0.53 & 0.24\\
Histruct+ & 14.43 & 0.99 & 0.00 & 0.63 & 0.94 & 0.54 & 0.30\\
\midrule
Lead3 & 13.03 & 1.00 & 0.00 & 0.49 & 0.95 & 0.28 & 0.05 \\
Textrank & 11.06 & 0.96 & 0.00 & 0.46 & 0.93 & 0.91 & 0.46\\
Textrank (ST) & $\phantom{0}$8.92 & 0.93 & 0.02 & 0.44 & 0.93 & 1.07 & 0.58 \\
PacSum (tfidf) & 12.89 & 0.99 & 0.01 & 0.49 & 0.94 & 0.59 & 0.33 \\
PacSum (bert) & 13.98 & 1.00 & 0.00 & 0.49 & 0.95 & 0.31 & 0.13 \\
MI-unsup & 10.62 & 0.99 & 0.00 & 0.46 & 0.92 & 1.05  & 0.38\\
\bottomrule
 \end{tabular*}
 }
\caption{All metric scores and the human Overall score for the 16 extractive systems on the 100 CNN/DM testing examples. The score of a system is the average score of 100 examples. $\downarrow$ means the scores are the lower the better.} 
\label{table:system_scores}
\vspace{-8pt}
\end{table*}

\section{Faithfulness Metric Details}\label{sec:appendix_metrics}
We select the following representative metrics to assess whether they can help to detect unfaithful summaries for extractive summarization. Unless otherwise stated, we use the original code provided by the official repository.

\textbf{ROUGE} \cite{lin-2004-rouge} is not designed for faithfulness evaluation; instead, it is the most widely used content selection evaluation metric for summarization. Although it has been shown that ROUGE correlates poorly with the human judgment of faithfulness \cite{maynez-etal-2020-faithfulness}, we explore whether it still holds for the extractive case. We only report ROUGE-2-F1 because other variants share similar trends with it. we use the implementation from the Google research Github repo.\footnote{\url{https://github.com/google-research/google-research/tree/master/rouge}}

\textbf{FactCC} \cite{kryscinski-etal-2020-evaluating} is an entailment-based metric trained on a synthetic corpus consisting of source sentences as faithful summaries and perturbed source sentences as unfaithful ones. It means that FactCC inherently treats each source sentence as faithful.
During the evaluation, they take the average score for each summary sentence as the final score.

\textbf{DAE} \cite{goyal-durrett-2020-evaluating} is also entailment-based and evaluates whether each dependency arc in the summary is entailed by the document or not. 
The final score is the average of arc-level entailment labels. DAE is similarly trained by a synthetic dataset compiled from paraphrasing. Since dependency arcs are within sentences, DAE also can hardly detect discourse-level unfaithfulness issues. 

\textbf{QuestEval} \cite{scialom-etal-2021-questeval} is a F1 style QGQA metric for both faithfulness and content selection evaluations. It first generates questions from both the document and the summary. Then, it answers the questions derived from the summary using the document (i.e., precision) and answers the questions derived from the summary using the summary (i.e., recall). The final score is their harmonic mean (i.e., F1).
QuestEval theoretically can detect \emph{incorrect coreference} because QG considers the long context of the summary and the document. However, it may not be able to capture the other three types of errors. 

\textbf{BERTScore} \cite{zhang2020bertscore} is a general evaluation metric for text generation. It computes the token-level cosine similarities between two texts using BERT \cite{devlin-etal-2019-bert}. 
Some previous works \cite{pagnoni-etal-2021-understanding, fischer2021finding} have shown that its \emph{precision} score between the summary and the source (i.e., how much summary information is similar to that in the document) has a good correlation with the summary's faithfulness.
We hypothesize BERTScore is able to capture more general discourse-level errors because of the contextualized representations from BERT.

Table~\ref{table:system_scores} show the metric scores as well as the human Overall score of the 16 systems we study in this work. Scores are computed only on the 100 CNN/DM testing examples we use, and the system score is the average of example scores.

\section{\exteval{} Details}\label{sec:appendix_exteval}

For \textbf{\incomcoref{}}, the list of pronouns we use includes \emph{they}, \emph{she}, \emph{he}, \emph{it}, \emph{this}, \emph{that}, \emph{those}, \emph{these}, \emph{them}, \emph{her}, \emph{him}, \emph{their}, \emph{her}, \emph{his}, and the list of determiners includes \emph{the}, \emph{that}, \emph{this}, \emph{these}, \emph{those}, \emph{both}. This list only contains frequent terms that appear in our dataset, which is not exhaustive.

The list of linking terms for \textbf{\incomdisco{}} includes \emph{and}, \emph{so}, \emph{still}, \emph{also}, \emph{however}, \emph{but}, \emph{clearly}, \emph{meanwhile}, \emph{not only}, \emph{not just}, \emph{on one side}, \emph{on another}, \emph{then}, \emph{moreover}. Similarly, the list is not exhaustive, and we only keep frequent terms.

\begin{figure*}[ht]
\begin{center}
\small
\begin{tabularx}{\textwidth}{X}
\toprule 
\textbf{Document:} \\
\underline{(CNN) The California Public Utilities Commission on Thursday said it is ordering Pacific Gas \& Electric Co. to pay a record} \underline{\$1.6 billion penalty for unsafe operation of its gas transmission system, including the pipeline rupture that killed eight people} \underline{in San Bruno in September 2010.} \\
Most of the penalty amounts to forced spending on improving pipeline safety. Of the $1.6 billion, $850 million will go to "gas transmission pipeline safety infrastructure improvements," the commission said. \\
Another \$50 million will go toward "other remedies to enhance pipeline safety," according to the commission. "PG\&E failed to uphold the public's trust," commission President Michael Picker said. \\
"The CPUC failed to keep vigilant. Lives were lost. Numerous people were injured. Homes were destroyed. \\
We must do everything we can to ensure that nothing like this happens again." The company's chief executive officer said in a written statement that PG\&E is working to become the safest energy company in the United States. \\
"Since the 2010 explosion of our natural gas transmission pipeline in San Bruno, we have worked hard to do the right thing for the victims, their families and the community of San Bruno," Tony Earley said. \\
"We are deeply sorry for this tragic event, and we have dedicated ourselves to re-earning the trust of our customers and the communities we serve. The lessons of this tragic event will not be forgotten." \\
On September 9, 2010, a section of PG\&E pipeline exploded in San Bruno, killing eight people and injuring more than 50 others. \\
The blast destroyed 37 homes. PG\&E said it has paid more than \$500 million in claims to the victims and victims' families in San Bruno, which is just south of San Francisco. \\
The company also said it has already replaced more than 800 miles of pipe, installed new gas leak technology and implemented nine of 12 recommendations from the National Transportation Safety Board. \\
\underline{According to its website, PG\&E has 5.4 million electric customers and 4.3 million natural gas customers.} \\
The Los Angeles Times reported the previous record penalty was a \$146 million penalty against Southern California Edison Company in 2008 for falsifying customer and worker safety data. CNN's Jason Hanna contributed to this report. \\
\midrule
\textbf{Summary (\emph{incomplete coreference)}:} \\
(CNN) The California Public Utilities Commission on Thursday said it is ordering Pacific Gas \& Electric Co. to pay a record \$1.6 billion penalty for unsafe operation of its gas transmission system, including the pipeline rupture that killed eight people in San Bruno in September 2010. According to \textbf{its} website, PG\&E has 5.4 million electric customers and 4.3 million natural gas customers. \\
\bottomrule
\end{tabularx}
\end{center}
\caption{An example from CNN/DM~\cite{hermann2015teaching} testing set showing an \emph{incomplete coreference} error. The summary is generated by BERT+LSTM+PN+RL \cite{zhong-etal-2019-searching}. All extracted sentences are \underline{underlined} in the document. The word \textbf{its} in the summary is ambiguous. It can refer to PG\&E or California Public Utilities Commission. The correct coreference should be PG\&E  in the document.}
\label{fig:coreference-example}
\end{figure*}

\section{Additional Examples}
Figure~\ref{fig:coreference-example} and Figure~\ref{fig:discourse-example} show two additional examples of \emph{incomplete coreference} and \emph{incomplete disource} respectively. Figure~\ref{fig:misleading-example} shows a misleading information example. Figure~\ref{fig:post-correction-example} is an example of fixing an incorrect coreference error via post-editing.

\begin{figure*}[ht]
\begin{center}
\small
\begin{tabularx}{\textwidth}{X}
\toprule 
\textbf{Document:} \\
(CNN) It's been a busy few weeks for multiples. \\
\underline{The first set of female quintuplets in the world since 1969} \underline{was born in Houston on April 8,} and the parents are blogging about their unique experience. \\
\underline{Danielle Busby delivered all five girls at the Woman's Hospital of Texas via C-section at 28 weeks and two days,} according to CNN affiliate KPRC. Parents Danielle and Adam and big sister Blayke are now a family of eight. \\
The babies are named Ava Lane, Hazel Grace, Olivia Marie, Parker Kate and Riley Paige. "We are so thankful and blessed," said Danielle Busby, who had intrauterine insemination to get pregnant. \\
"I honestly give all the credit to my God. I am so thankful for this wonderful hospital and team of people here. They truly all are amazing." You can learn all about their journey at their blog, "It's a Buzz World." \\
Early news reports said \underline{the Busby girls were the first all-female quintuplets} born in the U.S. \\
But a user alerted CNN to news clippings that show quintuplet girls were born in 1959 to Charles and Cecilia Hannan in San Antonio. \\
All of the girls died within 24 hours. Like the Busby family, Sharon and Korey Rademacher were hoping for a second child. \\
When they found out what they were having, they decided to keep it a secret from family and friends. \\
That's why they didn't tell their family the gender of baby No. 2 -- or that Sharon was actually expecting not one but two girls, according to CNN affiliate WEAR. \\
And when everyone arrived at West Florida Hospital in Pensacola, Florida, after Sharon gave birth March 11, they recorded everyone's reactions to meeting twins Mary Ann Grace and Brianna Faith. \\
The video was uploaded to YouTube on Saturday and has been viewed more than 700,000 times. Could you keep it a secret? \\
\midrule
\textbf{Summary (\emph{incomplete discourse)}:} \\
The first set of female quintuplets in the world since 1969 \\
was born in Houston on April 8, \\
Danielle Busby delivered all five girls at the Woman's Hospital of Texas via C-section at 28 weeks and two days, \\
\textbf{the Busby girls were the first all-female quintuplets} \\
\bottomrule
\end{tabularx}
\end{center}
\caption{An example from CNN/DM~\cite{hermann2015teaching} testing set showing an \emph{incomplete discourse} error. The summary is generated by the Oracle (disco) \cite{xu-etal-2020-discourse} extractive system. All extracted elementary discourse units are \underline{underlined} in the document. The last summary sentence missed the ``born in the u.s'' part which may make people think the Busby girls is the first all-female quintuplets not only in US.}
\label{fig:discourse-example}
\end{figure*}

\begin{figure*}[ht]
\begin{center}
\small
\begin{tabularx}{\textwidth}{X}
\toprule 
\textbf{Document:} \\
(CNN) It didn't seem like a fair fight. \\
\underline{On one side were hulking football players and pro wrestlers, competing as teams of two to eat as many pounds of steak as} \underline{they could, combined, in one hour.} \\
On another was a lone 124-pound mother of four. \\
\underline{And sure enough, in the end, Sunday's contest at Big Texan Steak Ranch in Amarillo, Texas, wasn't even close.} \\
Molly Schuyler scarfed down three 72-ounce steaks, three baked potatoes, three side salads, three rolls and three shrimp cocktails -- far outpacing her heftier rivals. \\ \underline{That's more than 13 pounds of steak, not counting the sides.} \\
And she did it all in 20 minutes, setting a record in the process. \\
"We've been doing this contest since 1960, and in all that time we've never had anybody come in to actually eat that many steaks at one time," Bobby Lee, who co-owns the Big Texan, told CNN affiliate KVII. "So this is a first for us, and after 55 years of it, it's a big deal." \\
In fairness, Schuyler isn't your typical 124-pound person. The Nebraska native, 35, is a professional on the competitive-eating circuit and once gobbled 363 chicken wings in 30 minutes. \\
Wearing shades and a black hoodie, Schuyler beat four other teams on Sunday, including pairs of football players and pro wrestlers and two married competitive eaters. \\
She also broke her own Big Texan record of two 72-ounce steaks and sides, set last year, when she bested previous record-holder Joey "Jaws" Chestnut. \\
...\\
\midrule
\textbf{Summary (\emph{other misleading information)}:} \\
On one side were hulking football players and pro wrestlers, competing as teams of two to eat as many pounds of steak as they could, combined, in one hour. \\
And sure enough, in the end, Sunday's contest at Big Texan Steak Ranch in Amarillo, Texas, wasn't even close. \\
That's more than 13 pounds of steak, not counting the sides. \\
\bottomrule
\end{tabularx}
\end{center}
\caption{An example from CNN/DM~\cite{hermann2015teaching} testing set showing a \emph{other misleading information} error. The summary is generated by the HeterGraph \cite{wang-etal-2020-heterogeneous} extractive system. All extracted sentences are \underline{underlined} in the document. If readers only read the summary, they may think the football players and pro wrestlers won the contest and ate 13 pounds of steak.}
\label{fig:misleading-example}
\end{figure*}

\begin{figure*}[ht]
\begin{center}
\small
\begin{tabularx}{\textwidth}{X}
\toprule 
\textbf{Document:} \\
\underline{(CNN) North Korea accused Mexico of illegally holding one of its cargo ships Wednesday and demanded the release of} \underline{the vessel and crew.} \\
\underline{The ship, the Mu Du Bong, was detained after it ran aground off the coast of Mexico in July.} \\
Mexico defended the move Wednesday, saying it followed proper protocol because the company that owns the ship, North Korea's Ocean Maritime Management company, has skirted United Nations sanctions. \\
... \\
But An Myong Hun, North Korea's deputy ambassador to the United Nations, said there was no reason to hold the Mu Du Bong and accused Mexico of violating \textcolor{blue}{the crew members'} human rights by keeping them from their families. \\
"Mu Du Bong is a peaceful, merchant ship and it has not shipped any items prohibited by international laws or regulations," An told reporters at the United Nations headquarters Wednesday. 
"And we have already paid full compensation to Mexican authorities according to its domestic laws." \\
According to Mexico's U.N. mission, the 33 North Korean nationals who make up \textcolor{blue}{the vessel's crew} are free, staying at a hotel in the port city of Tuxpan and regularly visiting the ship to check on it. \\
\underline{\textcolor{blue}{They} will soon be sent back to North Korea with help from the country's embassy, Mexican authorities said.} \\
In the case of the Chong Chon Gang, Panamanian authorities found it was carrying undeclared weaponry from Cuba -- including MiG fighter jets, anti-aircraft systems and explosives -- buried under thousands of bags of sugar. \\
Panama seized the cargo and held onto the ship and its crew for months. North Korea eventually agreed to pay a fine of \$666,666 for the vessel's release. CNN's Jethro Mullen contributed to this report. \\
\midrule
\textbf{Original Summary (\emph{incorrect coreference)}:} \\
(CNN) North Korea accused Mexico of illegally holding one of its cargo ships Wednesday and demanded the release of \textcolor{red}{the vessel and crew}. \\
The ship, the Mu Du Bong, was detained after it ran aground off the coast of Mexico in July. \\
\textcolor{red}{\bf They} will soon be sent back to North Korea with help from the country's embassy, Mexican authorities said. \\
\midrule
\textbf{Automatically Corrected Summary:} \\
(CNN) North Korea accused Mexico of illegally holding one of its cargo ships Wednesday and demanded the release of the vessel and crew. \\
The ship, the Mu Du Bong, was detained after it ran aground off the coast of Mexico in July. \\
\textcolor{blue}{the crew members'} will soon be sent back to North Korea with help from the country's embassy, Mexican authorities said. \\
\bottomrule
\end{tabularx}
\end{center}
\caption{An example of post-correction with \exteval{}. In the original summary, \emph{they} refers to \emph{the vessel and crew} in the summary, but it only refers to \emph{the crew} in the document. In the corrected summary, the automated program successfully replaces \emph{they} with \emph{the crew members'} though with a minor grammar issue.}
\label{fig:post-correction-example}
\end{figure*}

\end{document}